# ProgressiveSpinalNet architecture for FC layers

Praveen Chopra,
IIT Jodhpur, India

**Abstract**

In deeplearning models the FC (fully connected) layer has biggest important role for classification of the input based on the learned features from previous layers. The FC layers has highest numbers of parameters and fine-tuning these large numbers of parameters, consumes most of the computational resources, so in this paper it is aimed to reduce these large numbers of parameters significantly with improved performance. The motivation is inspired from SpinalNet and other biological architecture. The proposed architecture has a gradient highway between input to output layers and this solves the problem of diminishing gradient in deep networks. In this all the layers receives the input from previous layers as well as the CNN layer output and this way all layers contribute in decision making with last layer. This approach has improved classification performance over the SpinalNet architecture and has SOTA performance on many datasets such as Caltech101, KMNIST, QMNIST and EMNIST. The source code is available at *"https://github.com/praveenchopra/ProgressiveSpinalNet"*

## 1 INTRODUCTION

DeepLearning has given state of the art performance in various scientific and engineering fields, but it requires very large computational power and with lot of time for training. With advancement in the DeepLearning, more and more deep networks are being developed and training these deep network requires very large size of training dataset. These deep networks has millions of features and the final FC layers also becomes very big due to large size of input. The FC layers contains most of the parameters and plays most significant role in the classification performance. In a typical model the learned features from previous convolution layers are feed to these FC layers and the classification is made by these FC layers. In a typical structure the size of the first FC equals to CNN output and gradually size is increased and then gradually decreased to make final FC layer size equal to number of classes.

This FC architecture has been improved by Kabir[1] with SpinalNet, where all the input layers are connected to output layers and all layers contribute in the classification at final FC layer. This architecture was inspired with biological Spinal architecture where the human spinal cord receives senses of touch from different locations in different parts of it [2]. Then these inputs are given to brain for decision making.

In biological network the human Spinal does some processing of the inputs and then the processed features are given to the brain for final action [2]. With the same inspiration this ProgressiveSpinalNet architecture is designed. In this architecture, the input is progressively processed at each stage instead of merging all layers output at last FC stage. The final classification at last FC layer is based on the progressively processed inputs of previous layers. In this architecture final stage of FC layer receives the processed features from previous layers with the actual features of the CNN, so the classification performance is better than the SpinlNet and it has given state of the art performance (SOTA) in many of the datasets and in other datasets it has comparable performance. This enhancement in the performance is also due to the elimination of the vanishing gradient problem in the DNN's. In this architecture there is a gradient super highway from output to input such as in LSTM. This FC architecture can be easily plugged in various Deeplearning models with changing the size of input and output. Due to progressive architecture this FC network is called as ProgressiveSpinalNet.

## 2 Proposed architecture

The proposed architecture has an input to output connection as shown in the figure 1 given below. The human spinal is also works similar way where all the sensory inputs are collected be super nerves from small nerves and then feed to brain. In this transportation of the information the processing of the information is also done as reported in literatures by many biological theory's[6] [7][8]. The working of the spinal can be understood by from R. D'mello [2] theory that the Pain signal transmitted through the spinal are amplified at various stages and some pre-processing also done, so that if a pathway is not working, the information can be transmitted by other one. This is the basic working principle of our ProgressiveSpinalNet architecture.

In the ProgressiveSpinalNet architecture, the final layer of FC network receive the processed or extracted features in a hierarchical manner. So finally layer receives the both features extracted from CNN layers as well as features learned from previous layers of FC network. So, the decision making at final layer of FC is better than only using the previous FC layer FC layer features.

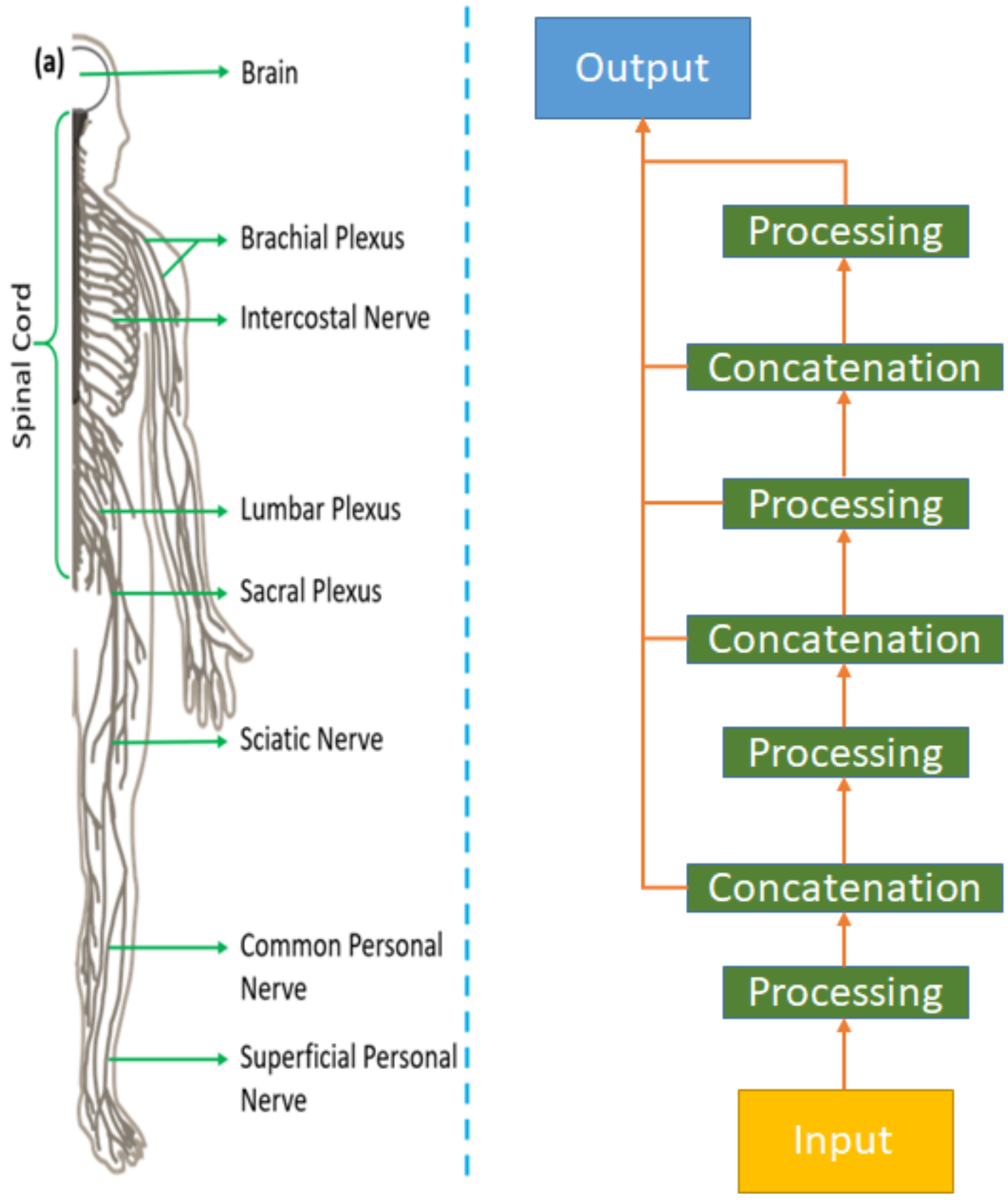

*Figure 1 Human spinal and ProgressiveSpinalNet architecture*

This concept is similar to FractalNet [5] as given in figure 2, which uses the similar type of processing in CNN layers. A. Byerl [3] and D. Ciregan [4] also proposed similar types of architectures for various CNN based networks for feature extraction. In our approach, instead of training with a single branch in an iteration, the ProgressiveSpinalNet train the whole FC network in a single iteration. The gradients are propagated back from last layer of FC to directly each layer of the FC due to connectivity available from first layer of FC to last layer of the FC. This way the network eliminates the problem of the vanishing gradients. This path from last layer to first layer of the FC works as the gradient super highway like LSTM.

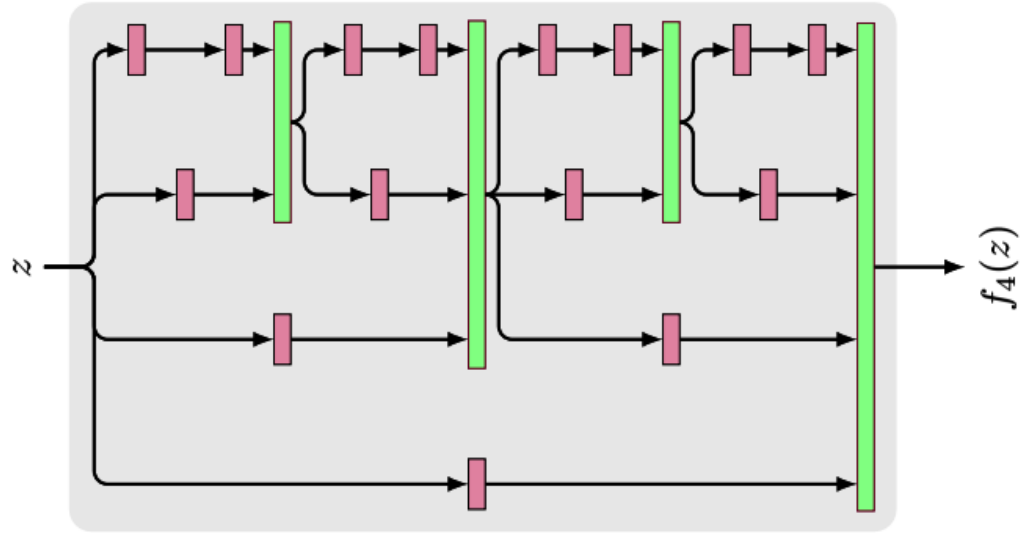

*Figure 2 FractalNet architecture*

The size of each layer is progressively increased in each layer of the FC. This gradual increase does not makes the network very bulky as the increment is equal to size of the first layer output. This much amount of neuron are added into each step of the FC network. In the proposed ProgressiveSpinalNet architecture, the output size is kept same for each FC layer. This made the network design simple and more effectively manageable. The figure 3 shows a typical ProgressiveSpinalNet architecture with input size and size of different layers of FC layers.

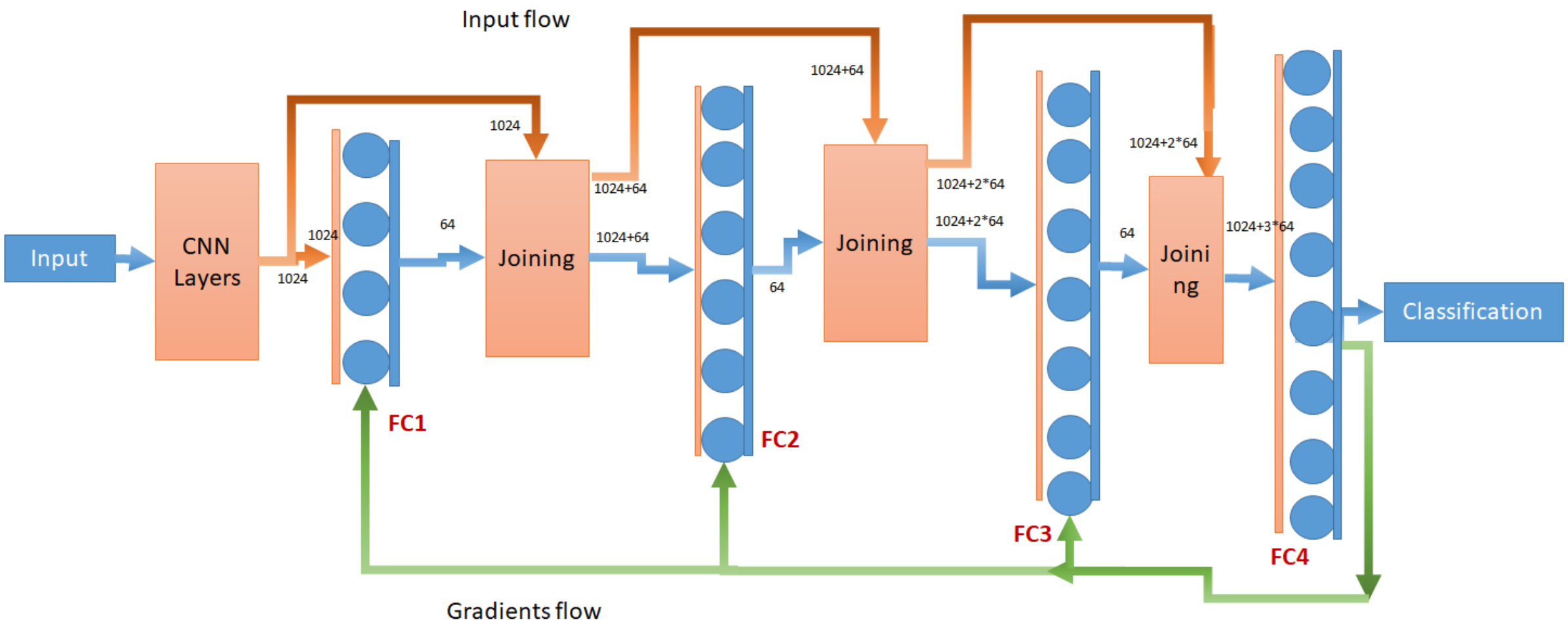

*Figure 3 ProgressiveSpinalNet FC connections gradient flow*

The general network of FC is a simple DNN network and has limitation on its depth, as deeper we go, the lesser the accuracy we achieves due to vanishing gradients problem. So the normal FC networks are mostly limited to depth of the 3-4 layers max. But the ProgressiveSpinalNet, does not suffers from the problem of the vanishing gradients so it depth can be much more than the normal FC network. This FC network has been tested with 6 layers and the performance was at par or better than the 3 layer normal FC network.

## 3 Results

The performance of this ProgressiveSpinalNet architecture has been checked for various datasets and it has performed very well. In most of the datasets it has given SOTA performance when it has used as FC layer of many standard pre-trained networks. If it is used on dataset where no pre-trained network is used, the performance is either at par or better than the normal FC network. In different datasets, as per the size of classification labels the output size is increased or decreased as per the size of ProgressiveSpinalNet and tested for many configuration. Its performance was always better than normal FC network.

For the performance comparison the SpinalNet is used as the reference because SpinalNet has SOTA on many data set [reference to paperwithcode.com] and baseline for this work. The performance of the ProgressiveSpinalNet is tested on various vanilla models along with the transfer learning models. The performance of the ProgressiveSpinalNet on both of these types is better than the SpinalNet or at par with SpinalNet.

This ProgressiveSpinalNet has been tested on various datasets and the performance with different hyper parameters is as given in table below:

| S. No | Dataset | Hyper parameters | Base Model | Accuracy (best of all epoch) in % | Remarks |
|---|---|---|---|---|---|
| 1. | MNIST | ✓ Learning rate=0.001, momentum=0.5<br>✓ Batch size=100<br>✓ HiddenLayer size = 128<br>✓ No. of epochs=50<br>✓ Negative Log-Likelihood Loss<br>✓ SGD with momentum = 0.9<br>✓ Dropout =0.2<br>✓ No of Layers =6 | Vanilla FC layer only, no CNN layers | 98.19 | Performance at par with reported results |
| 2. | EMNIST digits | ✓ Learning rate=0.005<br>✓ Batch size=50<br>✓ HiddenLayer size = 64<br>✓ No. of epochs=50<br>✓ CrossEntropy<br>✓ SGD with momentum = 0.9<br>✓ Dropout =0.2 | VGG-5 | 99.82 | **SOTA performance** |
| 1. | EMNIST Letters | ✓ Learning rate = 0.005<br>✓ Batch size = 50<br>✓ Hidden layer size = 128<br>✓ No. of epochs = 100<br>✓ CrossEntropy<br>✓ ADAM<br>✓ Dropout = 0.2 | VGG-5 | 95.8605 | Performance at par with reported results |
| 2. | QMNITS | ✓ Learning rate = 0.0008<br>✓ Batch size = 64<br>✓ Hiddenlayer size = 64 | VGG-5 | 99.6867 | **SOTA performance** |
| | | ✓ No. of epochs = 100<br>✓ CrossEntropy<br>✓ ADAM<br>✓ Dropout = 0.2 | | | |
| 3. | KMNITS | ✓ Learning rate = 0.0008<br>✓ Batch size = 64<br>✓ Hidden layer size = 64<br>✓ No. of epochs = 200<br>✓ CrossEntropy<br>✓ ADAM<br>✓ Dropout = 0.2 | VGG-5 | 98.98 | Performance at par with reported results |
| 4. | Bird225 | ✓ Learningrate=0.0008, momentum=0.9<br>✓ Batch size=16<br>✓ Hidden layer size = 256<br>✓ No of epochs=10<br>✓ CrossEntropyLoss<br>✓ SGD<br>✓ StepLR<br>✓ Dropout =0.5 | Pre trained wide_resnet101_2 Model | 99.55 | **SOTA performance** |
| 5. | Fruits 360 | ✓ Learningrate=0.0008momentum=0.9<br>✓ Batch size=16<br>✓ Hidden layer size = 256<br>✓ No of epochs=10<br>✓ CrossEntropyLoss<br>✓ SGD<br>✓ StepLR<br>✓ Dropout =0.5 | Pre trained wide_resnet101_2 | 99.97 | Performance at par with reported results |
| 6. | STL-10 | ✓ Learningrate=0.0008, momentum=0.9<br>✓ Batch size=8<br>✓ Hidden layer size = 256<br>✓ No of epochs=50<br>✓ CrossEntropyLoss<br>✓ SGD<br>✓ StepLR<br>✓ Dropout =0.5 | Pre trained wide_resnet101_2 Model | 98.18 | **SOTA performance** |
| 7. | Caltech101 | ✓ Learningrate=0.001, momentum=0.9<br>✓ Batch size=16<br>✓ HiddenLayer size = 200<br>✓ No of epochs=16<br>✓ CrossEntropyLoss<br>✓ SGD | Pre trained wide_resnet101_2 | 97.76 | **SOTA performance** |
| | | ✓ StepLR<br>✓ Dropout =0.3 | | | |

# 4 Conclusion:

In this paper an improved concept of DNN is presented with progressive computational network called ProgressiveSpinalNet for FC layers of deep-networks. This is a biologically inspired network design and proven its performance on various datasets with SOTA performance on many datasets. This network does not have problem of vanishing gradients with support of large depth of network. The structure of this network does not requires too many parameters, this leads to improvement in performance with less computations. This network concept can be easily expanded to wide range of real-world scenarios.